\documentclass[runningheads]{llncs}
\usepackage{amsmath}
\usepackage{amssymb}
\usepackage{booktabs}
\usepackage{graphicx}
\usepackage{xcolor}
\usepackage{hyperref}
\usepackage{array}

\newcolumntype{C}[1]{>{\centering\arraybackslash}p{#1}}
\begin{document}

\title{Deep learning based geometric registration for medical images: How accurate can we get without visual features?}
\titlerunning{Deep learning based geometric registration for medical images}
\author{Lasse Hansen \and Mattias P. Heinrich}
\authorrunning{L. Hansen and M. P. Heinrich}
\institute{Institute of Medical Informatics, Universität zu Lübeck, Lübeck, Germany}
\maketitle              % typeset the header of the contribution
\begin{abstract}
As in other areas of medical image analysis, e.g. semantic segmentation, deep learning is currently driving the development of new approaches for image registration. Multi-scale encoder-decoder network architectures achieve state-of-the-art accuracy on tasks such as intra-patient alignment of abdominal CT or brain MRI registration, especially when additional supervision, such as anatomical labels, is available. The success of these methods relies to a large extent on the outstanding ability of deep CNNs to extract descriptive visual features from the input images. In contrast to conventional methods, the explicit inclusion of geometric information plays only a minor role, if at all. In this work we take a look at an exactly opposite approach by investigating a deep learning framework for registration based solely on geometric features and optimisation. We combine graph convolutions with loopy belief message passing to enable highly accurate 3D point cloud registration. Our experimental validation is conducted on complex key-point graphs of inner lung structures, strongly outperforming dense encoder-decoder networks and other point set registration methods. Our code is publicly available at \url{https://github.com/multimodallearning/deep-geo-reg}.

\end{abstract}

\section{Introduction}
Current learning approaches for medical image analysis predominantly consider the processing of volumetric scans as a dense voxel-based task. However, the underlying anatomy could in many cases be modelled more efficiently using only a sparse subset of relevant geometric keypoints. When sufficient amounts of labelled training data are available and the region of interest can be robustly initialised, sparse surface segmentation models have been largely outperformed by dense fully-convolutional  networks in the past few years \cite{isensee2019automated}. However, dense learning based image registration has not yet reached the accuracy of conventional methods for the estimation of large deformations where geometry matters - e.g. for inspiration-expiration lung CT alignment. The combination of iconic (image-based) and geometric registration approaches have excelled in deformable lung registration but they are often time-consuming and rely on multiple steps of pre-alignment, mask-registration, graph-based optimisation and multi-level continuous refinement with different image-based metrics \cite{ruhaak2017estimation}. 

In this work, we aim to address 3D lung registration as a purely geometric alignment of two point clouds (a few thousand 3D points for inhale and exhale lungs each). While this certainly reduces the complexity of the dense deformable 3D registration task, it may also reduce the accuracy since intensity- and edge-based clues are no longer present. Yet, we demonstrate in our experimental validation that even this limited search range for potential displacements leads to huge and significant gains compared to dense learning based registration frameworks - mainly stemming from the robustness of our framework to implicitly learn the geometric alignment of vessel and airway trees.

\subsection{Related Work}
\textbf{Point Cloud Learning:} Conventional point cloud registration (iterative closest point, coherent point drift)\cite{myronenko2010point} often focused on the direct alignment of unstructured 3D points based on their coordinates. Newer work on graph convolutional learning has demonstrated that relevant geometric features can be extracted from point clouds with neighbourhood relations defined on kNN graphs and enable semantic labeling or global classification of shapes, object parts and human poses and gestures \cite{bronstein2017geometric,qi2017pointnet}. %The challenge to define an operation that works on data without regular grids and neighbourhoods, which still follows the principles of shared weights and locality can either be addressed with graph convolutions \cite{duvenaud2015convolutional} or the permutation invariant PointNet \cite{qi2017pointnet} and its extensions. A substantial amount of papers has discussed the semantic labelling or global classification of shapes, object parts and human poses and gestures using geometric deep learning. 
Graph Convolutional Networks (GCN) \cite{kipf2017semi} define localised filter and use a polynomial series of the graph Laplacian (Tschebyscheff polynomials) further simplified to the immediate neighbourhood of each node.% GCNs lead to an intuitive generalisation of convolution to unstructured (non-grid) graphs with weighted adjacency matrices, learns features that act only on the individual nodes (weights sharing across the graph) and propagates information along edges.
%An important limitation of graph convolution networks is their inability to dynamically learn the importance of edges in the adjacency or Laplace matrix.
The graph attention networks introduced in \cite{velivckovic2018graph} are a promising extension based on attention mechanism. Similarly, dynamic edge convolutions \cite{wang2019dynamic} achieve information propagation by learning a function that predicts pairwise edge weights based on previous features of both considered nodes. %that was popularised in machine translation and image analysis \cite{schlemper2019attention}. A related approach computes so called

\textbf{Learning Based Image Registration:} In image registration, learning based methods have surpassed their untrained optimisation-based counterparts in terms of accuracy and speed for 2D optical flow estimation, where millions of realistic ground truth displacement fields can be generated \cite{sun2018pwc}. Advantages have also been found for certain 3D medical registration tasks, for which thousands of scans with pixel-level expert annotations are available and the complexity of deformations is well represented in the training dataset \cite{balakrishnan2019voxelmorph,xu2019deepatlas,mok2020large}. As evident from a recent medical registration challenge \cite{hering_alessa_2020_3835682}, deep learning has not yet reached the accuracy and robustness for inspiration to expiration CT lung registration, where detailed anatomical labels are scarce (learning lobe alignment might not directly translate into low registration errors \cite{hering2020constraining}) and the motion is large and complex. Even for the simpler case of shallow breathing in 4DCT, few learning-based works have come close to the best conventional methods (e.g. \cite{ruhaak2017estimation}) despite increasingly complex network pipelines \cite{chen2020semantic,fu2020lungregnet}. 

\textbf{Learning Graphical Registration:} More recent research in computer vision has also explored geometric learning for 3D scene flow \cite{liu2019flownet3d} that aims to register two 3D point clouds by finding soft correspondences. The challenge stems from the difficulty of jointly embedding two irregular point cloud (sub-)sets to enable end-to-end learning of geometric features and correspondence scores. Other recent approaches in point set registration/matching combine deep feature learning with GCNs and classical optimisation techniques, to solve the optimal transport \cite{puy2020flot} or reformulate traditional matching algorithms into deep network modules \cite{sarlin2020superglue}. In the medical domain, combining sparse MRF-based registration \cite{sotiras2010simultaneous} and multi-level continuous refinement \cite{ruhaak2017estimation} yielded the highest accuracy for two 3D lung benchmarks comprising inspiration and expiration \cite{castillo2013reference,murphy2011evaluation}. 

We strongly believe that geometry can be a key element in advancing learning based registration and that the focus on visual features and fully-convolutional networks has for certain applications diverted research from mathematically proven graphical concepts that can excel within geometric networks. 

\subsection{Contribution}
We propose a novel geometric learning method for large motion estimation across lung respiration that combines graph convolutional networks on keypoint clouds with sparse message passing. Our method considers geometric registration as soft correspondence search between two keypoint clouds with a restricted set of candidates from the moving point cloud for each fixed keypoint. 
\textbf{1)} We are the first to combine edge convolutions as end-to-end geometric feature learning from sparse keypoints with differentiable loopy belief propagation (discrete optimisation) for regularisation of displacements on a kNN graph adapted to irregular sets of candidates for each node. 
\textbf{2)} Our compact yet elegant networks, demonstrate surprisingly large gains in accuracy and outperform deep learning approaches that make use of additional visual clues by more than 50\% reduced target registration errors for lung scans of COPD patients. 
\textbf{3)} We present a further novel variant of our approach that discretises the sparse correspondence probabilities using differentiable extrapolation for a further six fold gain in computational efficiency and with similar accuracy.

\section{Methods}

\begin{figure}[t]
\includegraphics[width=\textwidth]{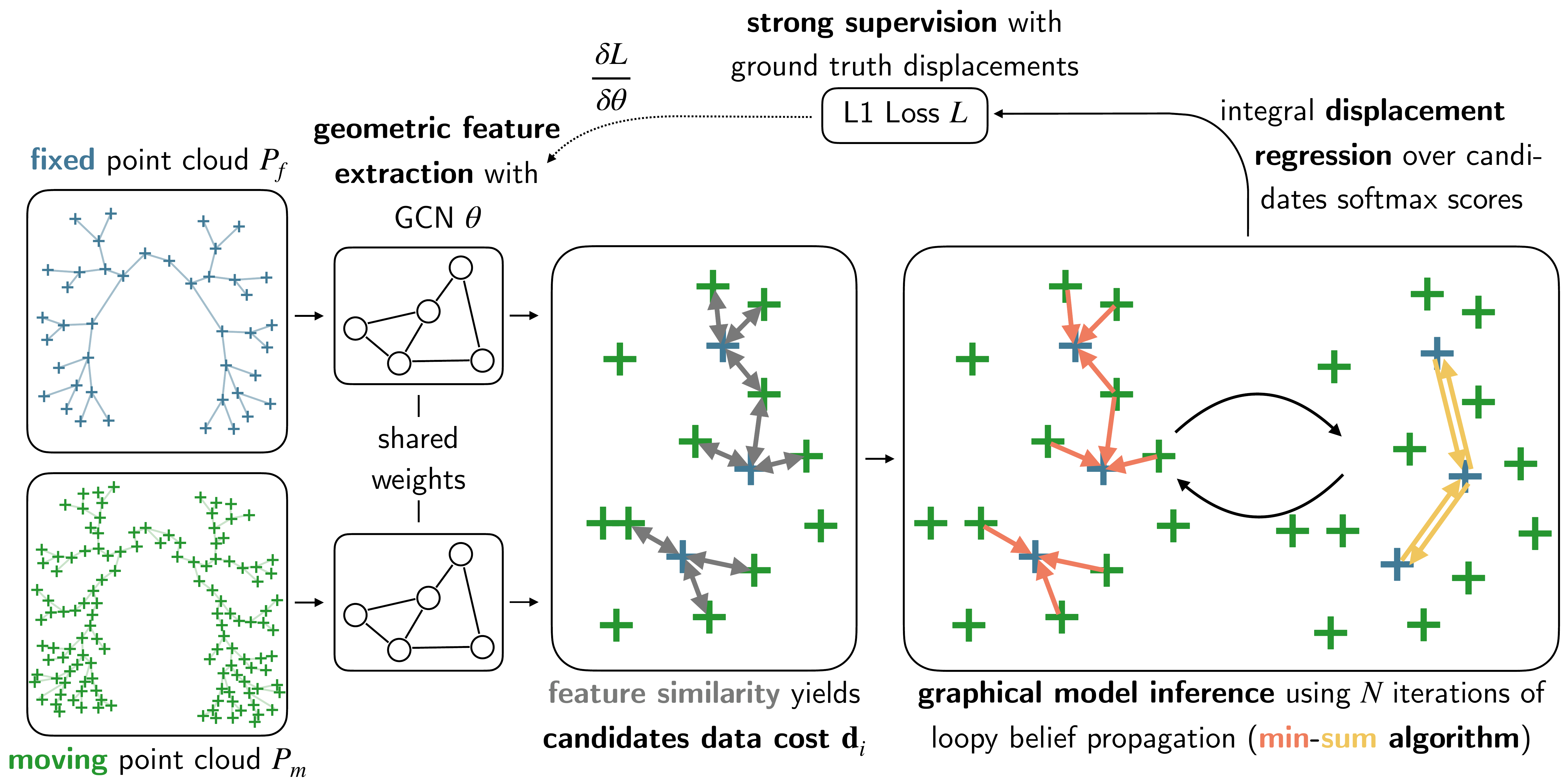}
\caption{Overview of our proposed method for accurate point cloud alignment using geometric features combined with loopy belief propagation in an end-to-end trainable deep learning registration framework.} \label{fig:overview}
\end{figure}

%In this section we first give a general description of our proposed graphical inference model for keypoint based registration and introduce a fast approximation of the approach. Second, we describe graph convolutional networks to extract geometric features from an input point cloud. Third, we summarize the full end-to-end registration framework and discuss architectural and training details.

\subsection{Loopy Belief Propagation for Regularised Registration of Keypoint Graphs}
\label{sec:lbp}

\begin{figure}[t]
\includegraphics[width=\textwidth]{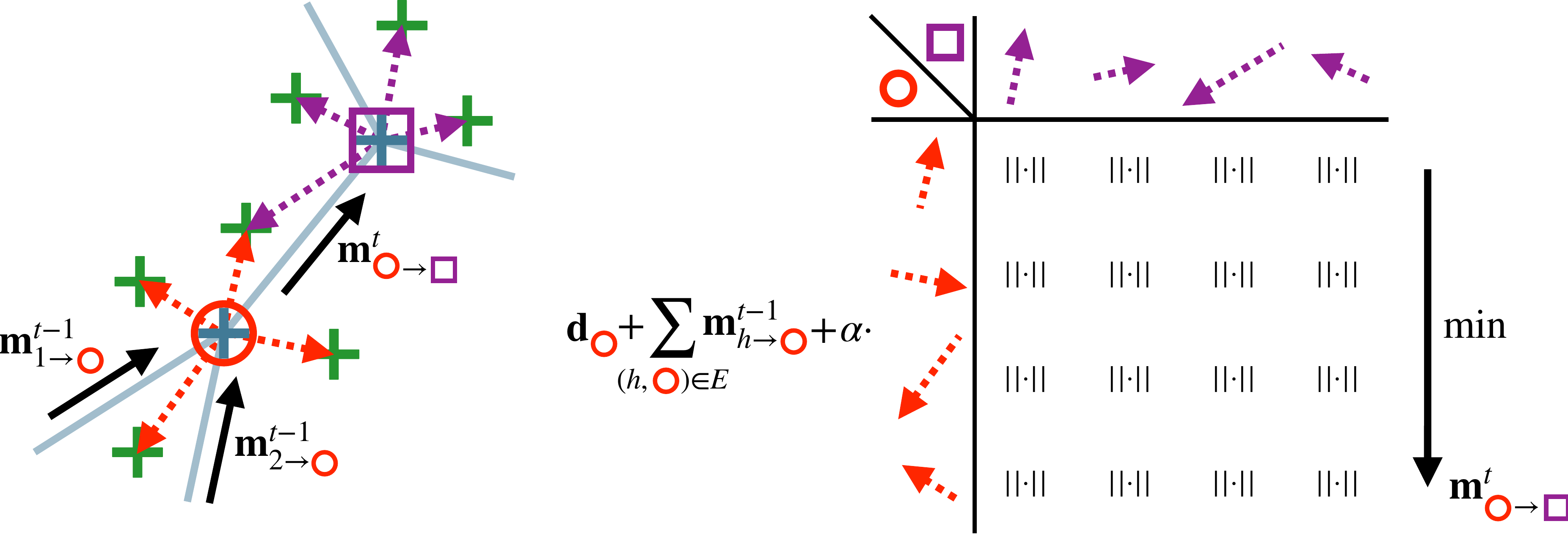}
\caption{Illustration of proposed message passing scheme for keypoint registration. The current outgoing message for the considered keypoint is composed of the candidates data cost and incoming messages from the previous iteration. In addition, the squared deviation (weighted by $\alpha$) of candidate displacements is minimised for a coherent motion across the $k$NN graph. Reverse messages are not shown for visual clarity.} \label{fig:lbp_details}
\end{figure}

We aim to align two point clouds, a fixed point cloud $P_f$ ($\vert P_f \vert = N_f$) and a moving point cloud $P_m$ ($\vert P_m \vert = N_m$). They consist of distinctive keypoints $\mathbf{p}_{f_i} \in P_f$ and $\mathbf{p}_{m_i} \in P_m$. We further define a symmetric $k$-nearest neighbour ($k$NN) graph on $P_f$ with edges $(ij) \in E$ that connect keypoints  $\mathbf{p}_{f_i}$ and $\mathbf{p}_{f_j}$. A displacement vector $\mathbf{v}_i \in V$ for each fixed keypoint $\mathbf{p}_{f_i}$ is derived from soft correspondences from a restricted set of possible candidates $\mathbf{c}^p_i \in C_i$ (determined by $l$-nearest neighbour search ($\vert C_i \vert = l$) in the moving point cloud $P_m$). The regularised motion vector field $V$ is inferred using loopy belief propagation enforcing spatial coherence of motion vectors. The data cost $d^p_i$ ($\mathbf{d}_i = (d^1_i,\dots,d^p_i,\dots,d^l_i)$) for a fixed point $\mathbf{p}_{f_i}$ and a single candidate $\mathbf{c}^p_i$ is modeled as
\begin{equation}
    d^p_i = \left\lVert\theta(\mathbf{p}_{f_i}) - \theta(\mathbf{c}^p_i)\right\rVert_2^2,
    \label{eq:data_cost}
\end{equation}
where $\theta(.)$ denotes a general feature transformation of the input point (e.g. deep learning based geometric features, cf. Section \ref{sec:gcn}). Especially in this case of sparse to sparse inference, missing or noisy correspondences can lead to sever registration errors. Therefore, a robust regularisation between neighbouring fixed keypoints (defined by edges $(ij) \in E$) is enforced by penalizing the deviation of relative displacements. The regularisation cost $r^{pq}_{ij}$ ($\mathbf{r}^{q}_{ij} = (r^{1q}_{ij},\dots,r^{pq}_{ij}, \dots,r^{lq}_{ij})$) for two fixed keypoints $\mathbf{p}_{f_i}, \mathbf{p}_{f_j}$ and candidates $\mathbf{c}^p_i, \mathbf{c}^q_j$ can then be described as
\begin{equation}
    r^{pq}_{ij}= \left\lVert(\mathbf{c}^p_i - \mathbf{p}_{f_i}) - (\mathbf{c}^q_j - \mathbf{p}_{f_j}) \right\rVert_2^2.
\end{equation}
To compute the marginal distributions of soft correspondences over the fixed $k$NN graph we employ $N$ iterations of loopy belief propagation (min-sum algorithm) with outgoing messages $\mathbf{m}^t_{i\rightarrow j}$ from $\mathbf{p}_{f_i}$ to $\mathbf{p}_{f_j}$ at iteration $t$ defined as
\begin{equation}
    \mathbf{m}^t_{i\rightarrow j} = \min_{1,\dots,q,\dots l}\bigg(\mathbf{d}_i + \alpha\mathbf{r}^q_{ij} - \mathbf{m}^{t-1}_{j\rightarrow i} + \sum_{(h,i) \in E}\mathbf{m}^{t-1}_{h\rightarrow i}\bigg).
\end{equation}
The hyperparameter $\alpha$ weights the displacement deviation penalty and thus controls the smoothness of the motion vector field $V$. Initial messages $\mathbf{m}^0_{i\rightarrow j}$ are set to $0$. A graphical description of the presented message passing scheme is also shown in Figure \ref{fig:lbp_details}.

\textbf{Fast Approximation Using a Discretised Candidates Space:}
While the proposed message passing approach is easily parallelisable, it still lacks some efficiency as the number of messages to compute for each keypoint is dependent on the number of neighbours $k$. We propose to reduce the number of message computations per node to $1$ by discretising the sparse candidates cost $\mathbf{d}_i$ in a dense cost volume $D_i$ with fixed grid resolution $r$. Voxelisation of sparse input has been used in point cloud learning to speed up computation \cite{liu2019point}. $D_i$ can be efficiently populated using nearest neighbour interpolation at (normalised) relative displacement locations $ \mathbf{o^p_i} = (o^p_{i_x},o^p_{i_y},o^p_{i_z}) = \mathbf{c}^p_i - \mathbf{p}_{f_i}$, evaluating
\begin{equation}
    D_i(u,v,w) = \frac{1}{N_{u,v,w}} \sum_{p=1}^l \mathbb{I}\big[
    \lfloor{o^p_{i_x}r} \rfloor = u,
    \lfloor{o^p_{i_y}r} \rfloor = v,
    \lfloor{o^p_{i_z}r} \rfloor = w \big] d^p_i,
\end{equation}
where (following notations in \cite{liu2019point}) $\mathbb{I}[\cdot]$ denotes a binary indicator that specifies whether the location $ \mathbf{o^p_i}$ belongs to the voxel grid $(u,v,w)$ and $N_{u,v,w}$ is a normalisation factor (in case multiple displacements end up in the same voxel grid). By operating on the dense displacement space $D_i$, we can employ an efficient quadratic diffusion regularisation using min convolutions \cite{felzenszwalb2006efficient} that are separable in dimensions and also avoid the costly computation of $k$ different messages per node. Approximation errors stem solely from the discretisation step.
%sampling the values of the lower envelope of parabolas (with heights equalling to the sum of the data cost and the previous message at iteration $t-1$) at the fixed displacement locations
\subsection{Geometric Feature Extraction with Graph Convolutional Neural Networks}
\label{sec:gcn}
Distinctive keypoint graphs that describe plausible shapes contain inherent geometric information. These include local features such as curvature but also global semantics of the graph (e.g. surface or structure connectivity). Recent work on data-driven graph convolutional learning has shown that descriptive geometric features can be extracted from point clouds with neighbourhood relations defined based on $k$NN graphs. Edge convolutions \cite{wang2019dynamic} can be interpreted as irregular equivalents to dense convolutional kernels. Following notations in \cite{wang2019dynamic} we define edge features $\mathbf{e}_{ij}=h_{\theta}(\mathbf{f}_i, \mathbf{f}_j - \mathbf{f}_i)$, where $\mathbf{f}_i$ denote $F$-dimensionsal features on points $\mathbf{p}_i \in P$ (first feature layer given as $\mathbf{f}_i = \mathbf{p}_i$). The edge function $h_{\theta}$ computes the Euclidean inner product of the learnable parameters $\theta=(\theta_1,\dots,\theta_F')$ with $\mathbf{f}_i$ (keypoint information) and $\mathbf{f}_j - \mathbf{f}_i$ (local neigbourhood information). The $F'$-dimensional feature output $\mathbf{f'}_i$ of an edge convolution is then given by
\begin{equation}
\mathbf{f}^{'}_i = \max_{(i,j) \in E} \mathbf{e}_{ij},
\end{equation}
where the max operation is to be understood as a dimension-wise aggregation function. Employing multiple layers of edge convolutions in a graph neural network and applying it to the fixed and moving point clouds ($P_f$, $P_m$) yields descriptive geometric features, which can be directly used to compute candidate data costs (see Equation \ref{eq:data_cost}).

\subsection{Deep Learning Based End-to-End Geometric Registration Framework}
Having described the methodological details, we now summarise the full end-to-end registration framework (see Figure \ref{fig:overview} for an overview). Input to the registration framework are the fixed $P_f$ and moving $P_m$ point cloud. In a first step, descriptive geometric features are extracted from $P_f$ and $P_m$ with a graph convolutional network $\theta$ (shared weights). The network consists of three edge convolutional layers, whereby edge functions are implemented as three layers of $1\times1$ convolutions, instance normalisation and leaky ReLUs. Feature channels are increased from $3$ to $64$. Two $1\times1$ convolutions output the final $64$-dimensional point feature embeddings. Thus, the total number of free trainable parameters of the network is 26880. In general, the moving cloud will contain more points than the fixed cloud (to enable an accurate correspondence search). To account for this higher density of $P_m$, the GCN $\theta$ acts on the $k$NN graph for $P_f$ and on the $3k$NN graph for $P_m$. As described in Section \ref{sec:lbp} the geometric features $\theta(P_f)$ and $\theta(P_m)$ are used to compute the candidates cost and final marginal distributions are obtained from $N$ iterations of (sparse or discretised) loopy belief propagation. As all operations in our optimisation step are differentiable the network parameters can be trained end-to-end. The training is supervised with ground truth motion vectors $\mathbf{\hat{v}}_i \in \hat{V}$ using an L1 loss (details on integral regression of the predicted motion vector field $V$ from the marginals distribution in Section \ref{sec:details}). 

\subsection{Implementation Details: Keypoints, Visual Features and Integral Loss}
\label{sec:details}
While our method is generally applicable to a variety of point cloud tasks, we adapted parts of our implementation to keypoint registration of lung CT.

\textbf{Keypoints:} We extract F\"{o}rstner keypoints with non-maximum suppression as described in \cite{heinrich2015estimating}. A corner score (distinctiveness volume) is computed using $D(x)=1/\operatorname{trace}\left((G_{\sigma}\ast(\nabla F\nabla F^T))^{-1}\right)$, where $G_\sigma$ describes a Gaussian kernel and $\nabla F$ spatial gradients of the fixed/moving scans computed with a seven-point stencil. Additionally, we modify the extraction to allow for a higher spatial density of keypoints in the moving scan by means of trilinear upsampling of the volume before non-maximum suppression. Only points within the available lung masks are considered.   

\textbf{Visusal Features:} To enable a fair comparison to state-of-the-art methods that are based on image intensities, we also evaluate variants of all geometric registration approaches with local MIND-SSC features \cite{heinrich2013towards}. These use a $12$-channel representation of local self-similarity and are extracted as small patches of size $3\times 3\times 3$ with stride=$2$. The dimensionality is then further reduced from 324 to 64 using a PCA (computed on each scan pair independently). 

\textbf{Integral Loss:} As motivated before, we aim to find soft correspondences that enable the estimation of relative displacements, without directly matching a moving keypoint location, but rather a probability for each candidate. A softmax operator over all candidates is applied to the negated costs after loopy belief propagation (multiplied by a heuristic scalar factor). This is integrated over the corresponding relative displacements. When considering a discretised search space (the faster dLBP variant), final displacement vectors are obtained via integration over the fully quantised 3D displacement space. 

To obtain a dense displacement field for evaluation (landmarks do not necessarily coinciding with keypoints), all displacement vectors of the sparse keypoints are accumulated in a displacement field tensor using trilinear extrapolation and spatial smoothing. This differentiable dense extrapolation enables the use of an L1 loss on (arbitrary) ground truth correspondences.

\section{Experiments and Results}

\begin{table*}[t]
    \centering
    \caption{Results of methods based on geometric features and optimisation on the COPDgene dataset \cite{castillo2013reference}. We report the target registration error (TRE) in millimeters for individual cases as well as the average distance and standard deviation over all landmarks. The average GPU runtime in seconds is listed in the last row.}
    \begin{tabular}{C{1.5cm}|C{1.5cm}|C{1.5cm}C{1.5cm}C{1.5cm}|C{1.5cm}C{1.5cm}}
        \toprule
                    & init. & CPD   & CPD+GF   & sLBP & sLBP+GF (ours) & dLBP+GF (ours)    \\
        \midrule
        \# 01       & 26.33 & 3.02  & 2.75          & 2.55  & \textbf{1.88} & 2.14          \\
        \# 02       & 21.79 & 10.83 & \textbf{5.96} & 8.69  & 6.22          & 6.69          \\
        \# 03       & 12.64 & 1.94  & 1.88          & 1.56  & \textbf{1.53} & 1.68          \\
        \# 04       & 29.58 & 2.89  & 2.84          & 3.57  & \textbf{2.63} & 3.01          \\
        \# 05       & 30.08 & 3.01  & 2.70          & 3.01  & \textbf{2.02} & 2.42          \\
        \# 06       & 28.46 & 3.22  & 3.65          & 2.85  & \textbf{2.21} & 2.69          \\
        \# 07       & 21.60 & 2.52  & 2.44          & 1.87  & \textbf{1.64} & 1.83          \\
        \# 08       & 26.46 & 3.85  & 3.58          & 2.08  & \textbf{1.93} & 2.14          \\
        \# 09       & 14.86 & 2.83  & 2.58          & 1.53  & \textbf{1.55} & 1.82          \\
        \# 10       & 21.81 & 3.57  & 5.57          & 3.15  & \textbf{2.79} & 3.72          \\
        \midrule
        avg         & 23.36 & 3.77  & 3.40          & 3.08  & \textbf{2.44} & 2.81          \\
        std         & 11.86 & 2.54  & 1.35          & 2.09  & 1.40          & 1.50          \\
        time        &       & 7.63  & 7.66          & 2.91  & 3.05          & \textbf{0.49} \\
        \bottomrule
    \end{tabular}
    \label{tab:results_geo}
\end{table*}

To demonstrate the effectiveness of our novel learning-based geometric 3D registration method, we perform extensive experimental validation on the DIR-Lab COPDgene data \cite{castillo2013reference} that consists of 10 lung CT scan pairs at full inspiration (fixed) and full expiration (moving), annotated with 300 expert landmarks each. Our focus lies in evaluating point cloud registration without visual clues and we extract a limited number of keypoints (point clouds) in fixed ($\approx$2000 each) and moving scans ($\approx$6000 each) within the lungs. 
Since, learning benefits from a variability of data, we add 25 additional 3D scan pairs showing inhale-exhale CT from the EMPIRE10 \cite{murphy2011evaluation} challenge, for which no landmarks are publicly available and we only include automatic correspondences generated using \cite{heinrich2015estimating} for supervision. We performed leave-one-out cross validation on the 10 COPD scans with sparse-to-dense extrapolation for landmark evaluation. Training was performed with a batch size of 4 and an initial learning rate of 0.01 for 150 epochs. All additional hyperparamters for baselines and our proposed methods (regularisation cost weighting $\alpha$, scalar factor for integral loss, etc.) were tuned on case \#04 of the COPDgene dataset and left unaltered for the remaining folds.

\begin{figure}[t]
\centering
\includegraphics[width=\textwidth]{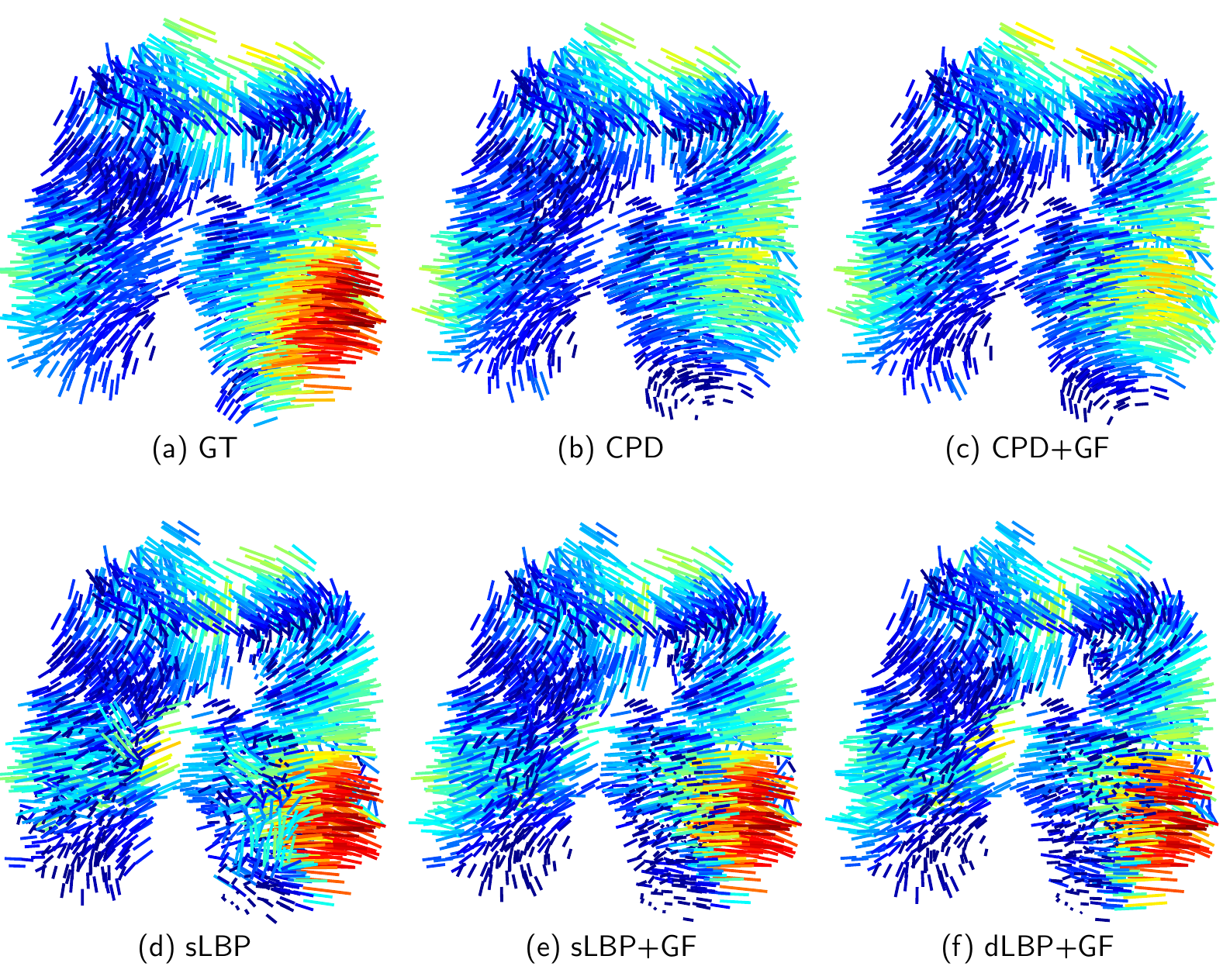}
\caption{Qualitative results of different geometric methods ((b)-(f)) on case \# 01 of the COPDgene dataset \cite{castillo2013reference}. The ground truth motion vector field is shown in (a). Different colors encode small (blue) and large motion (red).} \label{fig:results_qual}
\end{figure}

Overall, we compare five different algorithms that work purely on geometric information, five further methods that use visual input features and one deep-learning baseline for dense intensity registration (the winner of the Learn2Reg 2020 challenge LapIRN \cite{mok2020large}). Firstly, we compare our proposed sparse-LBP regularisation with geometric feature learning (sLBP+GF) to a version without geometric learning and coherent point drift \cite{myronenko2010point} without (CPD) and with geometric feature learning (CPD+GF). In addition, we evaluate the novel discretisation of sparse candidates that is again integrated into an end-to-end geometric learning with differentiable LBP regularisation (dLBP+GF) and leads to substantial efficiency gains. The results clearly demonstrate the great potential of keypoint based registration for the complex task of large deformable lung registration. Numerical and qualitative results are shown in Table \ref{tab:results_geo} and Figure \ref{fig:results_qual}, respectively. Even the baseline methods using no features at all, CPD and sLBP, where inference is based only on optimisation on the extracted keypoint graphs, achieve convincing target registration errors of $3.77~mm$ and $3.08~mm$. Adding learned geometric features within our proposed geometric registration framework leads to relative improvements of $10\%$ (CPD+GF) and $20\%$ (sLBP+GF), respectively. For the efficient approximation of our proposed appraoch (dLBP+GF) the TRE increases by approximate $0.35~mm$ but at the same time the average runtime is improved six fold to just below $0.5$ seconds (which is highly competitive with dense visual deep learning methods such as LapIRN (cf. Table \ref{tab:results_vis})). A statistical test (Wilcoxon signed-rank test calculated over all landmark pairs of the dataset) with respect to our proposed method (sLBP+GF) shows that improvements on all other comparison methods are highly significant ($p<0.001$).

\begin{table*}[t]
    \centering
    \caption{Results of methods based on visual features on the COPDgene dataset \cite{castillo2013reference}. We report the average target registration error (TRE) and standard deviation in millimeters over all landmarks. The average GPU runtime in seconds is listed in the last column. For an easier comparison we also add the results of our "geometry only" approaches.}
    \begin{tabular}{r|C{1.5cm}C{1.5cm}C{1.5cm}}
        \toprule
                            & avg           & std   & time          \\
        \midrule
        init{ }             & 23.36         & 11.86 &               \\
        \midrule
        FLOT+MIND{ }        & 5.87          & 1.30  & 1.63          \\
        LapIRN{ }           & 4.99          & 1.98  & 1.08          \\
        FE+sLBP+MIND{ }     & 3.83          & 1.21  & 16.71         \\
        sPDD+MIND{ }        & 3.16          & 0.69  & 2.17          \\
        CPD+MIND{ }         & 2.40          & 0.81  & 13.12         \\
        \midrule
        sLBP+MIND (ours){ } & \textbf{1.74} & 0.38  & 4.65          \\
        sLBP+GF (ours){ }   & 2.44          & 1.40  & 3.05          \\
        dLBP+GF (ours){ }   & 2.81          & 1.50  & \textbf{0.49} \\
        \bottomrule
    \end{tabular}
    \label{tab:results_vis}
\end{table*}

We made great efforts to use state-of-the-art learning-based 3D scene flow registration methods and obtained only meaningful results when incorporating the visual MIND features for FLOT \cite{puy2020flot} and heavily adapting the FlowNet3d embedding strategy \cite{liu2019flownet3d} (denoted as FE+sLBP+MIND). FlowNet3d aims to learn a flow embeddings (FE) using a concatenation of two candidate sets (from connected graph nodes), which does not lead to satisfactory results due to the permutation invariant nature of these sparse candidates. Hence, we designed a layer that captures all pairwise combinations and leads to a higher dimensional intermediate tensor that is fed into $1\times 1$ convolutions and is projected (with max-pooling) to a meaningful message vector. For FLOT, we replaced the feature extraction with the handcrafted MIND-PCA embeddings and also removed the refinement convolutions after the optimal transport block (we observed severe overfitting in our training setting when employing the refinement). The sPDD method is based on the probabilistic dense displacement (PDD) network and was modified to operate on the sparse fixed keypoints (instead of a regular grid as in the original published work \cite{heinrich2019closing}). Results for the state-of-the-art learning based 3D scene flow registration methods and further comparison experiments using visual input features can be found in Table \ref{tab:results_vis}. Our proposed sparse registration approach using visual MIND features (sLBP+MIND) achieves a TRE well below $2~mm$ and thus, improves on the geometry based equivalent (sLBP+GF) by $0.7~mm$. However, the extraction of visual features slows down the inference time by $1.6$ and $4.1$ (dLBP+GF) seconds, respectively. Notably, all proposed geometric registration methods achieve results on par with or significantly better (e.g. more than 50\% gain in target registration error w.r.t the dense multi-scale network LapIRN) than the deep learning based comparison methods with additional visual features.

\section{Discussion and Conclusion}
We believe our concept clearly demonstrates the advantages of decoupling feature extraction and optimisation by combining parallelisable differentiable message passing for sparse correspondence finding with graph convolutions for geometric feature learning. Our method enables effective graph-based pairwise regularisation and compact networks for robustly capturing geometric context for large deformation estimation. It is much more capable for 3D medical image registration as adaptations of scene flow approaches, which indicates that these methods may be primarily suited for aligning objects with repetitive semantic object/shape parts that are well represented in large training databases. 

We demonstrated that even without using visual features, the proposed geometric registration substantially outperforms very recent deep convolutional registration networks that excelled in other medical tasks. The reason for this large performance gap can firstly lie in the complexity of aligning locally ambiguous structures (vessels, airways) that undergo large deformations and that focusing on relatively few relevant 3D keypoints is a decisive factor in learning meaningful geometric transformations. Our new idea to discretise the sparse candidate displacements into a dense embedding using differentiable extrapolation yields immensive computational gains by reducing the number of message computations (from $k=9$ to 1 per node) and thereby also enabling future use within alternative regularisation algorithms.  

While our experimental analysis was so far restricted to lung anatomies, we strongly believe that graph-based regularisation models combined with geometric learning will play an important role for tackling other large motion estimation tasks, the alignment of anatomies across subjects for studying shape variations and tracking in image-guided interventions. Being able to work independently of visual features opens new possibilities for multimodal registration, where our method only requires comparable keypoints to be found, e.g. using probabilistic edge maps \cite{oktay2015structured}. In addition, the avoidance of highly parameterised CNNs can establish new concepts to gain a better interpretability of deep learning models.

\bibliographystyle{splncs04}
\bibliography{ipmi2021_sparse.bib}
\end{document}